\documentclass[10pt,twocolumn,letterpaper]{article}

\usepackage[pagenumbers]{cvpr} 

\usepackage{graphicx}
\usepackage{amsmath}
\usepackage{amssymb}
\usepackage{booktabs}
\usepackage{array}

\usepackage{array}
\newcolumntype{T}[1]{>{\raggedright\arraybackslash\let\newline\\\arraybackslash\vspace{0pt}}p{#1}}
\newcolumntype{C}[1]{>{\centering\arraybackslash\vspace{0pt}}p{#1}}

\usepackage[pagebackref,breaklinks,colorlinks]{hyperref}
\usepackage[misc]{ifsym}

\usepackage[capitalize]{cleveref}
\crefname{section}{Sec.}{Secs.}
\Crefname{section}{Section}{Sections}
\Crefname{table}{Table}{Tables}
\crefname{table}{Tab.}{Tabs.}

\def\cvprPaperID{*****} 
\def\confName{CVPR}
\def\confYear{2025}

\begin{document}

\title{Evaluation of Large Language Models for Anomaly Detection in Autonomous Vehicles}

\author{
Petros Loukas$^{1}$ \quad
David Bassir$^{2,3}$ \quad
Savvas Chatzichristofis$^{4}$ \quad
Angelos Amanatiadis$^{1}$\textsuperscript{\Letter} \\
$^{1}$ Democritus University of Thrace, Greece \\
$^{2}$ Dongguan University of Technology, China \\
$^{3}$ ENS-Paris-Saclay University, France \\
$^{4}$ Neapolis University Pafos, Cyprus \\
{\tt\small \{petrlouk1,aamanat\}@pme.duth.gr, david.bassir@ens-paris-saclay.fr, s.chatzichristofis@nup.ac.cy}
}
\maketitle

\begin{abstract}
   The rapid evolution of large language models (LLMs) has pushed their boundaries to many applications in various domains. Recently, the research community has started to evaluate their potential adoption in autonomous vehicles and especially as complementary modules in the perception and planning software stacks.  
   However, their evaluation is limited in synthetic datasets or manually driving datasets without the ground truth knowledge and more precisely, how the current perception and planning algorithms would perform in the cases under evaluation. 
   For this reason, this work evaluates LLMs on real-world edge cases where current autonomous vehicles have been proven to fail. The proposed architecture consists of an open vocabulary object detector coupled with prompt engineering and large language model contextual reasoning. 
   We evaluate several state-of-the-art models against real edge cases and provide qualitative comparison results along with a discussion on the findings for the potential application of LLMs as anomaly detectors in autonomous vehicles.
\end{abstract}

\let\thefootnote\relax\footnotetext{\Letter \: Corresponding author.}
\section{Introduction}
\label{sec:intro}

Autonomous Vehicles (AVs) have significantly advanced over the past decade, becoming increasingly proficient in handling typical driving tasks. Nonetheless, the deployment of AVs in dynamic and semi-structured driving scenarios continues to present significant challenges, primarily related to the safe handling of rare or unpredictable scenarios, commonly referred to as edge cases \cite{koopman2017autonomous,koopman2018autonomous}. Addressing these edge cases is crucial for ensuring safety and reliability, as they often involve unexpected elements or uncommon arrangements of usual scene constituents, thereby creating semantic anomalies that standard perception and planning systems fail to accurately interpret \cite{philion2020learning}. These edge cases are considered the main barrier for a wider adoption of autonomous vehicles \cite{moradloo2024safety}. 

Recently, Large Language Models (LLMs) have shown significant promise in a variety of applications beyond natural language processing, demonstrating strong capabilities in contextual reasoning, anomaly detection, and decision-making tasks \cite{brown2020language,wei2022chain}. Their ability to leverage extensive prior knowledge and reasoning skills has sparked interest in exploring their integration into AV systems to enhance perception and planning modules, particularly, but also in the identification and handling of semantic anomalies \cite{vemprala2024chatgpt}.

Semantic anomalies differ from traditional perception challenges \cite{zamanakos2021comprehensive,chalvatzaras2022survey,papadeas2021real} in that they involve familiar objects arranged or depicted in unusual contexts, making them particularly problematic for conventional machine learning models trained primarily on routine data distributions \cite{jeong2020ood}. Such anomalies demand advanced contextual reasoning capabilities that transcend simple pattern recognition, thus positioning LLMs as promising candidates to supplement existing AV architectures.

In this paper, we extend the current research by evaluating the role of LLMs as real-time contextual supervisors within both the perception and planning stacks of AVs. More precisely, we propose a modular architecture combining an open-vocabulary object detector with tailored prompt engineering to enable LLMs to assess scene semantics and flag anomalies in both nominal and edge-case scenarios. To this end, we evaluate multiple state-of-the-art LLMs on a curated set of real autonomous driving edge cases that have been classified as anomalies, and provide qualitative insights. Our results discuss the LLMs’ ability to generalize to novel contexts, revealing their potential as complementary reasoning agents in safety-critical autonomous driving systems.
\section{Related Work}

Recent advancements in autonomous vehicle (AV) technology have prompted extensive research into anomaly detection using advanced machine learning techniques. Language-enhanced latent representations, particularly through the use of models like  Contrastive Language Image Pre-training \cite{radford2021learning}, offer promising improvements by enabling out-of-distribution (OOD) scenarios to be defined explicitly through natural language inputs. This approach enhances both transparency and adaptability when an AV system encounters unforeseen or complex driving scenarios \cite{wang2023language}.

Vision Large Language Models (VLLMs) have further propelled anomaly detection methodologies by integrating visual data directly with natural language prompts. Such integration allows AVs to adapt to novel and unexpected obstacles more robustly, often achieving performance comparable to contemporary state-of-the-art models \cite{kim2023driving}. Similarly, rule-based reasoning frameworks utilize large language models to derive rules from regular patterns in videos, subsequently applying these rules to detect anomalies effectively. These techniques improve accuracy in identifying anomalous behaviors or situations, further contributing to the AV system's overall safety \cite{zhao2023follow}.

Semantic anomaly detection also benefits significantly from the integration of conventional large language models. By converting visual sensor data into natural language descriptions, current large language models can exploit contextual understanding to identify semantic anomalies, such as misinterpreted traffic signals, thereby enhancing reliability and decision-making capabilities in complex driving environments \cite{elhafsi2023semantic}. Real-time anomaly detection frameworks, leveraging both rapid binary classifiers and more comprehensive reasoning stages with large language models, can further enhance safety and reliability in AV systems, particularly when operating under strict resource constraints \cite{10771587}.

A closely related effort is the AESOP-LLM, an anomaly detection and reactive planning framework proposed by Sinha et al. \cite{sinha2024realtime}. Their system adopts a two-stage reasoning pipeline: i) a fast binary anomaly classifier that detects semantic anomalous conditions by querying similarity with previously recorded observations within the contextual embedding space of an LLM, and 
ii) a slower generative reasoner, assessing the safety implications of detected anomalies and selecting the appropriate mitigation strategy.   
This runtime monitoring framework utilizes generalist foundation models to facilitate safe and real-time control of AV systems faced with real-world anomalies. 

Finally, multimodal frameworks designed explicitly for challenging driving or environmental conditions, can significantly improve anomaly detection \cite{10716620}. All previous works have shown improved capability of reasoning, however, they are evaluated in synthetic or autonomous vehicles datasets where the ground-truth of existing perception and decision-making modules is not apparent.

\section{Real Edge Cases}

To assess the effectiveness of large language models in identifying semantic anomalies, it is essential to evaluate them against real-world edge cases that have challenged current autonomous driving systems. These are scenarios in which AVs failed not due to faulty sensors or wrong detection algorithms, but because the perception system misinterprets contextually complex or rare driving scenes. Unlike traditional out-of-distribution examples that differ significantly from training data at the pixel or feature level \cite{jeong2020ood}, semantic anomalies involve in-distribution elements arranged in unexpected or misleading ways, making them particularly difficult for conventional perception stacks to handle.\par
The dataset used in this study is hand-curated and consists of real-world edge cases that have been publicly reported or documented as having caused failures in autonomous vehicle systems, as shown in Table \ref{tab1}. These scenarios are characterized by their semantic irregularities that fall outside the typical distribution of driving environments encountered during training and validation. They include cases that, while visually plausible, exhibit atypical contextual combinations that mislead AV perception or decision-making modules.

\begin{table*}[t]
  \centering
  \footnotesize
  \begin{tabular}{T{0.8cm} C{2.5cm} T{10.5cm} T{1.5cm}}
    \toprule
    \textbf{Edge Case \#} & \textbf{Dashboard camera view} &
    \textbf{Anomaly description} & \textbf{Affected AV sub-system} \\
    \midrule
    1 & \includegraphics[width=\linewidth,height=1.5cm,keepaspectratio]{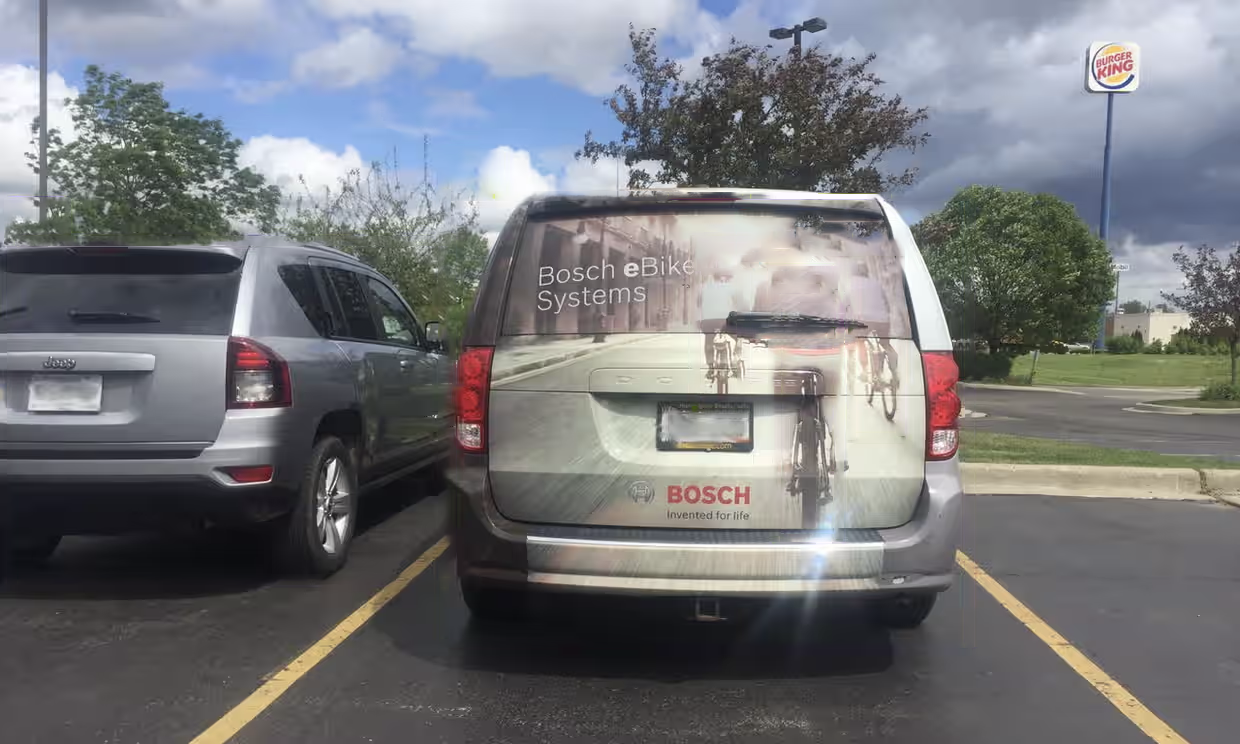} &
    A vehicle's rear window and trunk are covered with a high-resolution advertisement depicting cyclists riding on a road. This visual illusion can potentially mislead perception or planning systems into falsely identifying active bicyclists in the lane, triggering unnecessary evasive actions or false positive detections. &
    Perception, Planning \\
    \midrule
    2 & \includegraphics[width=\linewidth,height=1.5cm,keepaspectratio]{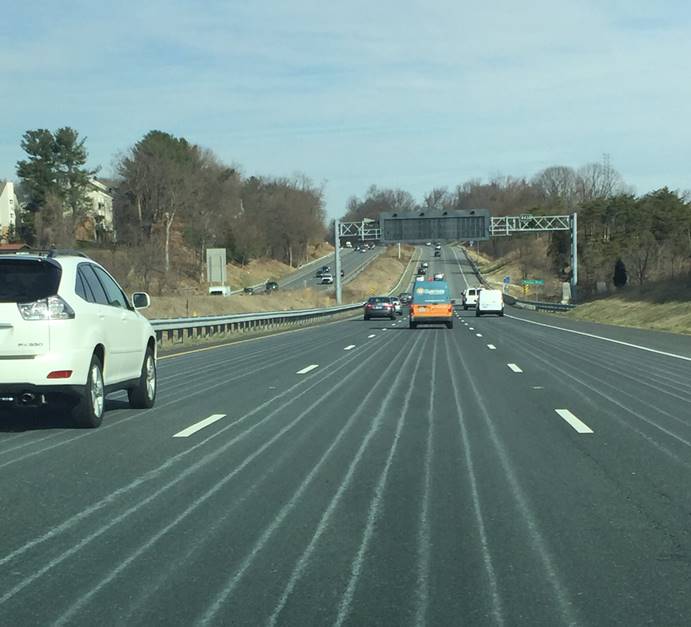} &
    Residual salt lines from a preceding truck create visual patterns on the road that resemble lane markings. These patterns may mislead lane detection algorithms, potentially causing incorrect lane localization or unsafe path planning decisions. &
    Perception, Planning \\
    \midrule
    3 & \includegraphics[width=\linewidth,height=1.5cm,keepaspectratio]{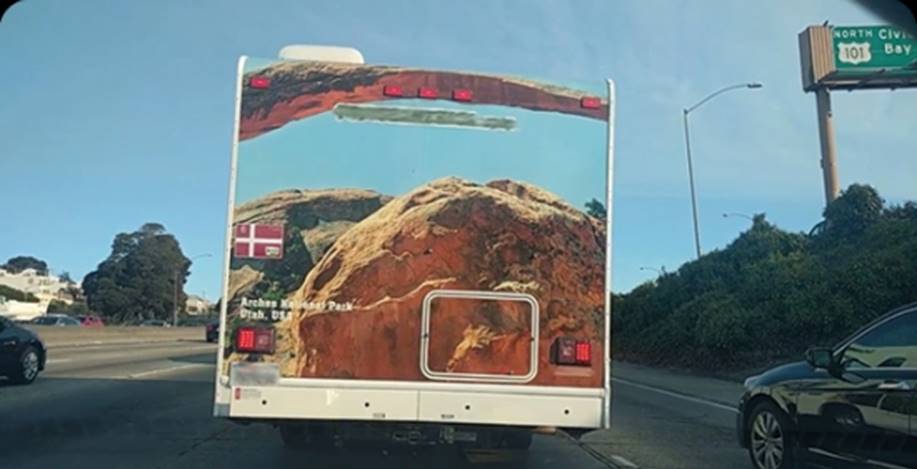} &
    The rear part of the vehicle is covered with a high-resolution poster depicting a mountainous landscape. Such visual overlays can confuse perception models, potentially causing misclassification of terrain, road boundaries, or environmental context. &
    Perception, Planning \\
    \midrule
    4 & \includegraphics[width=\linewidth,height=1.5cm,keepaspectratio]{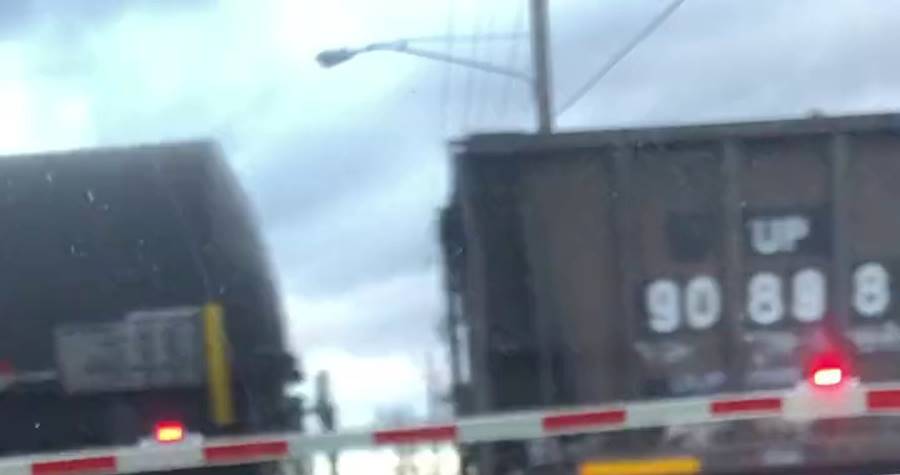} &
    Train wagons passing through a railway crossing were mistakenly identified by the perception system as buses. Such misclassification can lead to inappropriate behavioral responses, especially in scenarios requiring precise object tracking and motion prediction near intersections or crossings. &
    Perception \\
    \midrule
    5 & \includegraphics[width=\linewidth,height=1.5cm,keepaspectratio]{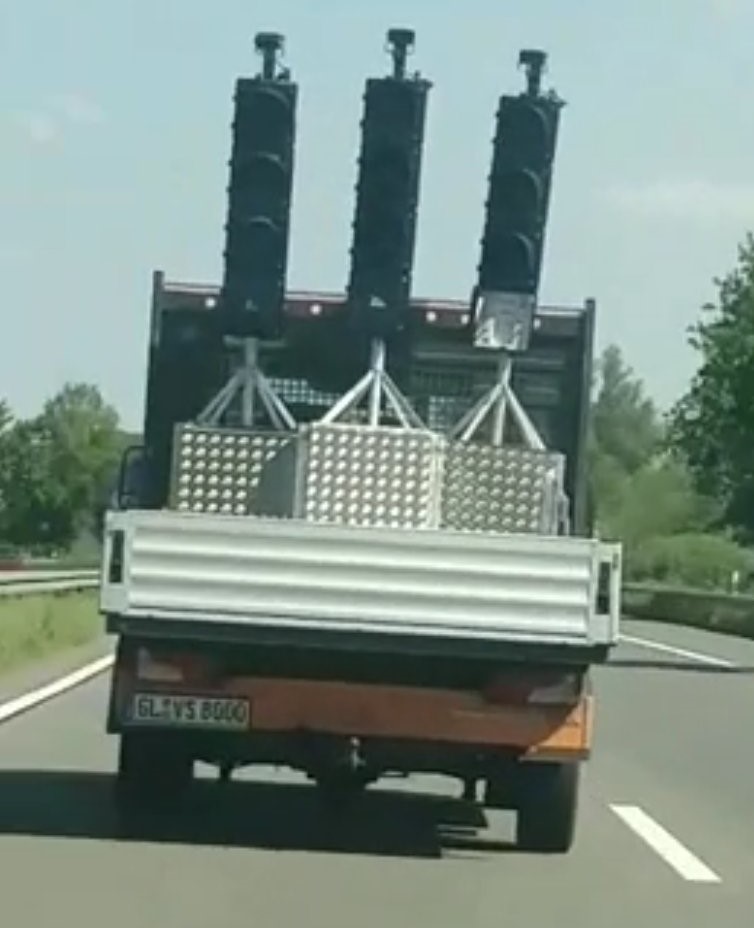} &
    A maintenance truck carrying inactive traffic lights. The configuration was misinterpreted by the autonomous perception system as functional roadside signals, leading to premature stops and incorrect traffic rule adherence. &
    Perception, Planning \\
    \midrule
    6 & \includegraphics[width=\linewidth,height=1.5cm,keepaspectratio]{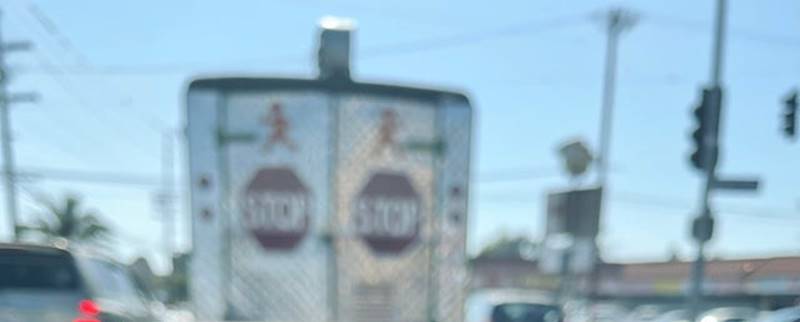} &
    A rear of a truck with printed graphics resembling stop signs. These visual patterns led to false detections by traffic sign recognition system, causing the autonomous vehicle to halt unnecessarily in flowing traffic. &
    Perception, Planning \\
    \midrule
    7 & \includegraphics[width=\linewidth,height=1.5cm,keepaspectratio]{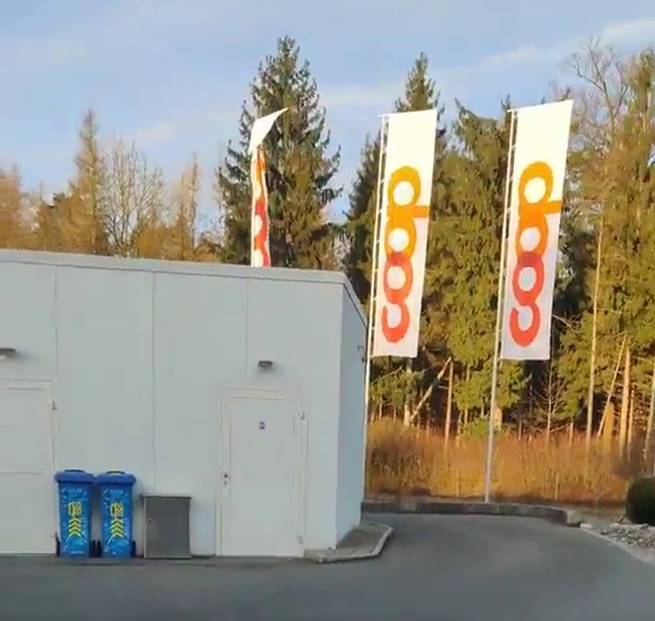} &
    Vertically waving flags with bold red and yellow elements were misclassified as traffic lights transitioning between red, yellow and green states. This led to erroneous decisions in behavior planning, including unnecessary stops. &
    Perception \\
    \midrule
    8 & \includegraphics[width=\linewidth,height=1.5cm,keepaspectratio]{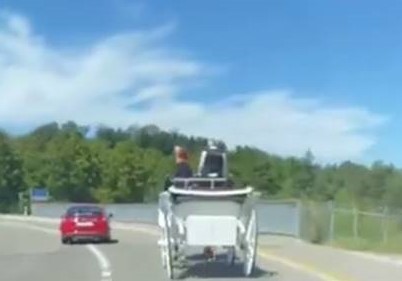} &
    A horse-drawn buckboard traveling on a public road alongside vehicles. Such uncommon and slow-moving road users may not be recognized or properly classified by autonomous perception systems, posing a challenge for safe interaction and appropriate behavioral planning. &
    Perception, Planning \\
    \midrule
    9 & \includegraphics[width=\linewidth,height=1.5cm,keepaspectratio]{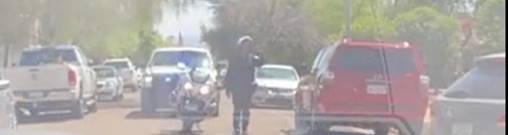} &
    A traffic officer is manually directing vehicles in an urban setting, while the autonomous vehicle comes to a complete halt. The AV system was unable to interpret the officer’s hand gestures, revealing a critical gap in dynamic, non-verbal human interaction understanding. &
    Planning \\
    \midrule
    10 & \includegraphics[width=\linewidth,height=1.5cm,keepaspectratio]{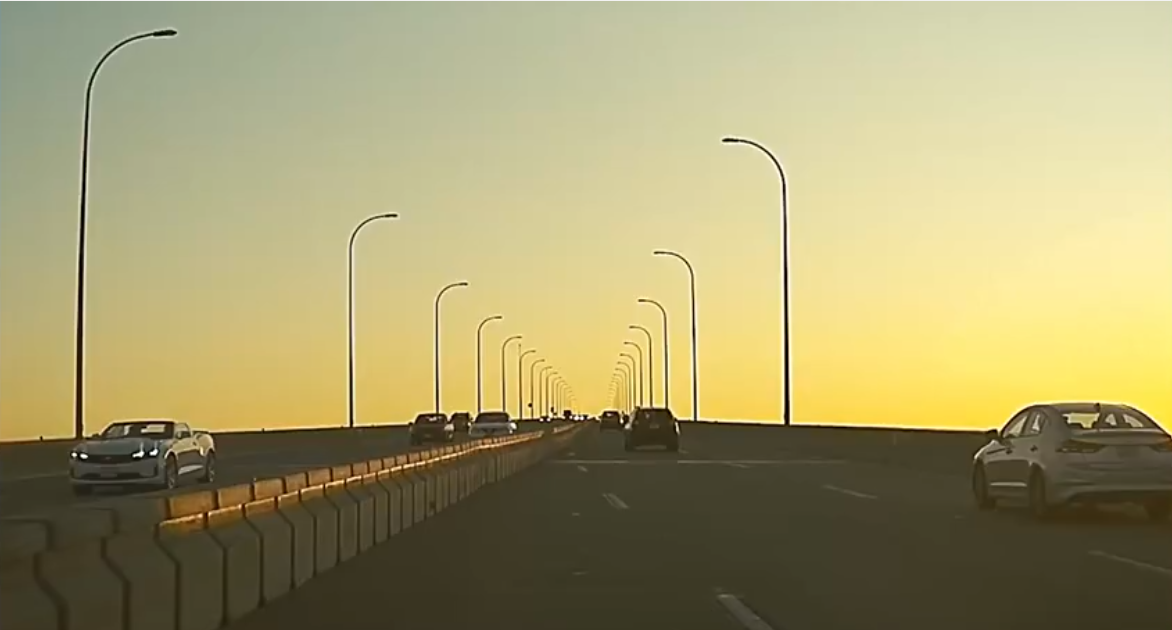} &
    Collision with the road parapet as it gradually encroached into the driving lane over an extended distance. The slow, progressive merge of the barrier was not correctly interpreted by the perception system, leading to a failure in lateral path adjustment and collision avoidance. &
    Perception \\
    \midrule
    11 & \includegraphics[width=\linewidth,height=1.5cm,keepaspectratio]{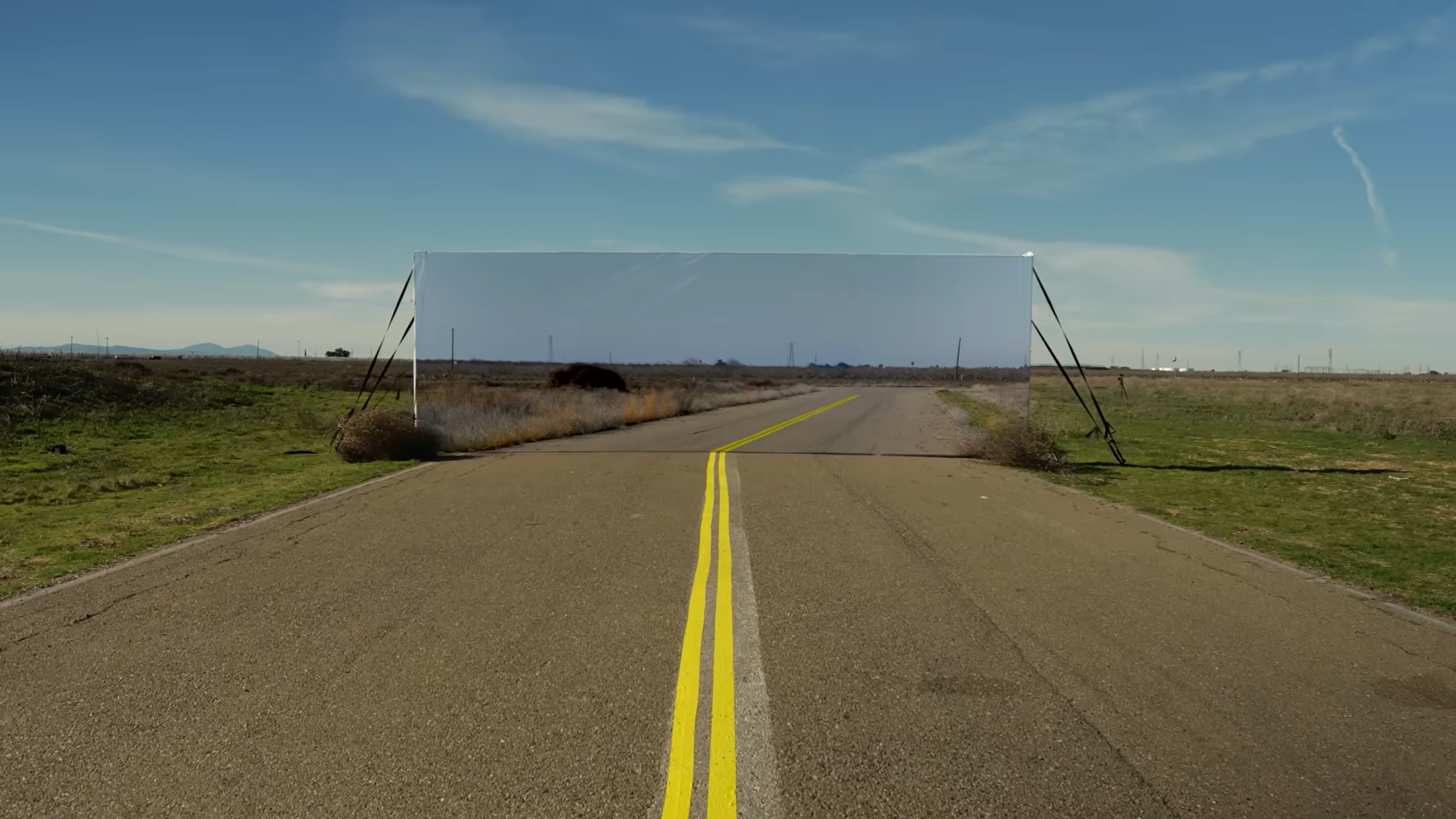} &
    A realistic-looking road backdrop printed on a large panel positioned across an actual roadway, creating the illusion that the road continues unobstructed. Visual deception that can critically mislead autonomous vehicle perception systems, potentially resulting in failure to stop or execute avoidance maneuvers in time. &
    Perception \\
    \midrule
    12 & \includegraphics[width=\linewidth,height=1.5cm,keepaspectratio]{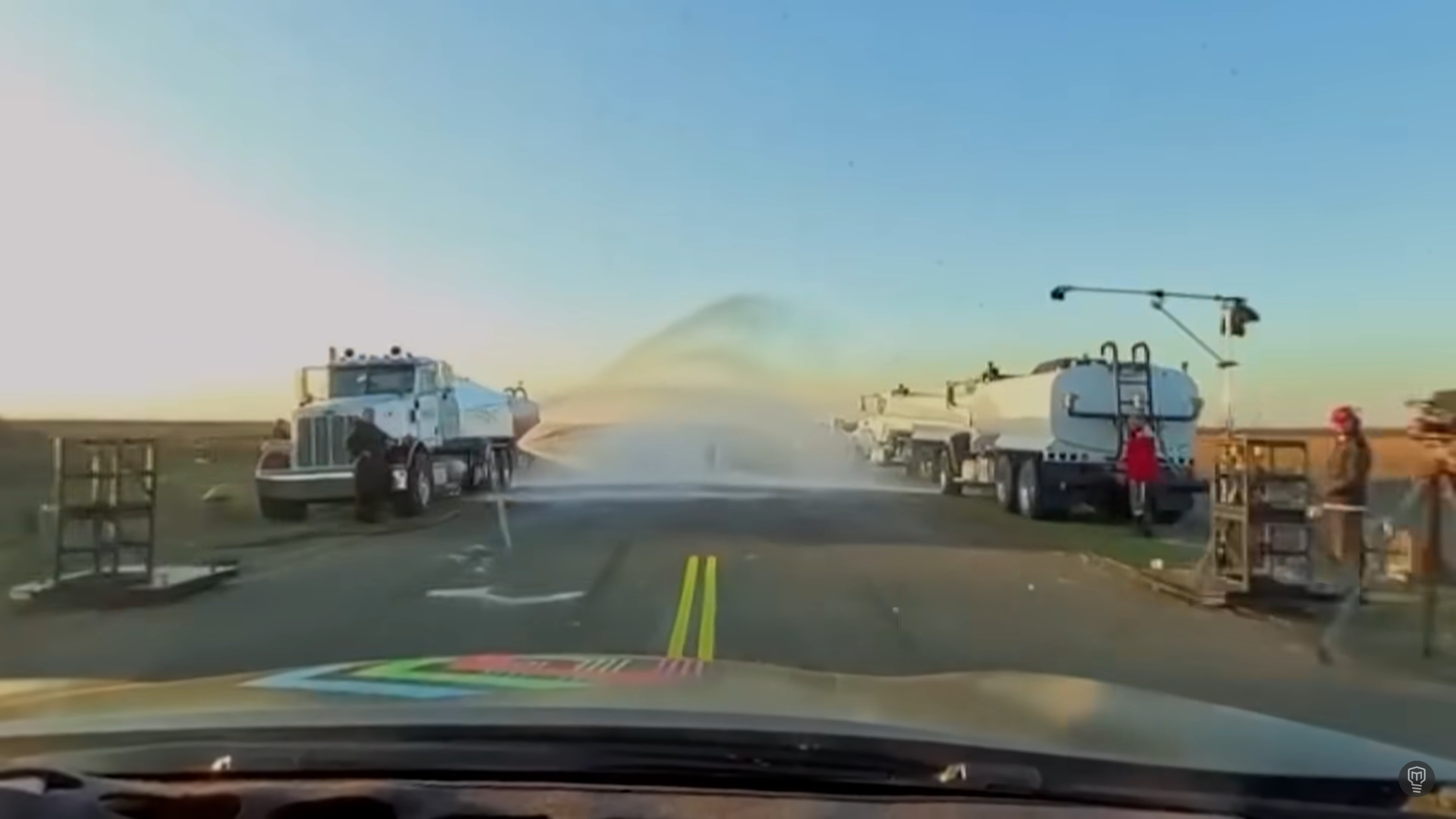} &
    A road being heavily sprayed with water, creating a dense spray curtain that obscures visibility. A pedestrian dummy standing behind the water spray is partially occluded, posing a significant risk as perception systems fails to detect its presence. &
    Perception \\
    \bottomrule
  \end{tabular}
  \caption{Real edge case dataset used for the LLM anomaly detection evaluation.}
  \label{tab1}
\end{table*}

These edge cases highlight essential challenges in semantic anomaly detection, particularly emphasizing the features involved in accurately interpreting dynamic scenes. Standard perception algorithms, primarily based on learned patterns from conventional datasets, are shown that have failed to contextualize the listed uncommon visual patterns or transient elements in the environment. For instance, the cases involving deceptive visual overlays like the mountainous poster on a truck or a road backdrop visually extending the actual road underscore the need for robust semantic reasoning beyond mere pixel-level analysis. Such examples underline that autonomous vehicle systems require additional layers of contextual understanding, enabling them to discern between genuine road elements and visually plausible yet deceptive driving situations.

\section{Proposed Methodology}
The selected edge cases were firstly parsed by an open vocabulary object detector to yield a scene description, consisting of the detected objects along with relevant context. The deployment of an open vocabulary object detector was essential for our methodology because semantic anomalies in autonomous vehicle edge cases often involve novel or rare objects and configurations that closed-set detector fail to capture. Unlike conventional models train on fixed sets (e.g. YOLO, Faster R-CNN), open-vocabulary detectors leverage language-vision alignment to recognize objects beyond the training dataset by using natural language queries \cite{minderer2022simple}. This flexibility is crucial when dealing with contextually deceptive scenarios such as trucks carrying traffic lights or misleading billboards where anomaly detection depends on identifying unexpected combinations of otherwise familiar elements.

Among open-vocabulary models, we specifically chose OWL-ViT to generate the scene description for each edge case. OWL-ViT is a zero-shot text-conditioned object detection model introduced by \cite{minderer2022simple}. It uses CLIP as its backbone to get multi-modal visual and text features. The image and text encoder are trained from scratch, with a contrastive loss, on the same image-text representations. To transfer the model to detection, \cite{minderer2022simple} took the pre-trained CLIP model and fine-tuned it end-to-end with the classification and box heads on standard detection datasets using a bipartite matching loss. Both the image and the text model are fine-tuned end-to-end. During the process of fine-tuning the final token pooling layer is removed, and instead a lightweight classification and box head is attached to each transformer output token. According to \cite{minderer2022simple}, open-vocabulary classification is enabled by replacing the fixed classification layer weights with the class-name embeddings obtained from the text model. The general detection training procedure of the model is almost identical to that for closed-vocabulary detectors. The difference is that OWL-ViT is not trained on fixed-label data, but on set of object category names which are provided as queries for each image. The classification head therefore outputs logits over the per-image label space defined by the queries, rather than a fixed global label space.

The reason we chose to deploy OWL-ViT, instead of another multimodal object detection model, is because it uniquely combines a simple yet powerful Vision Transformer backbone with large-scale contrastive pretraining and end-to-end fine-tuning. According to Minderer et al. \cite{minderer2022simple}, OWL-ViT supports zero-shot text-conditioned and one-shot image-conditioned detection, enabling robust generalization to unseen objects without additional retraining. Its modular design, where image and text encoders remain separate, allows efficient querying with thousands of object descriptions, making it particularly suited for large-scale anaomaly monitoring in AVs. Furthermore, OWL-ViT's ability to generate rich scene descriptions aligns with the requirements of our LLM-based reasoning pipeline, where detection quality directly influences the contextual understanding of anomalies.
At inference time, OWL-ViT can identify objects not explicitly present in the training dataset, thanks to the shared embedding space of the vision and text encoders. This capability makes it possible to use both images and textual queries for object detection, enhancing its versatility. To generate descriptive labels, we constructed a comprehensive vocabulary of visual concepts, compromising both normal road-scene elements and scenario-specific edge case anomalies. Each image was queried with a unified text prompt composed of two  text prompts per image to search for the target objects. One text prompt for the normal classes and one for the anomalous.

\subsection{Common road objects}
These represent the typical entities an autonomous vehicle may encounter and are grouped into three categories: 

• Road Vehicles: including \textit{Car, Truck, Police force, Ambulance, SUV, Box truck, Van, Motorcycle, Taxi}, and others.
    
• Infrastructure and Signage: including\textit{ Traffic light, Traffic sign, Street lamp, Billboard, Construction sign, Barrier, Sidewalk, Guide signage,} among others.
    
• Road Markings and Pedestrians: such as \textit{Pedestrian, Lane line, Broken white line, Solid yellow line, Center line,} etc. 

A common vocabulary was introduced to OWL-ViT, as listed in Table \ref{tab:my-table}.

\begin{table*}[t]
  \centering
  \small
  \begin{tabular}{p{4.5cm} p{11.5cm}}
    \toprule
    \textbf{Vocabulary type} & \textbf{Vocabulary objects} \\
    \midrule
    common\_road\_objects &
    Car, Truck, Bicycle, Semi-trailer truck, Motorcycle, SUV, Van, Wagon, Convertible, Coupe, Minivan, Scooter, Sedan, Four-wheeler, Garbage truck, Tractor, Taxi, Police force, Ambulance, Firetruck, Heavy truck, Box truck, Coach-Style bus, Passenger car, Pickup truck. \\
    \midrule
    infrastructure\_and\_signage &
    Building, House, Tree, Bin, Fire hydrant, Mailbox, Road tunnel, Overhead variable message sign, Parking sign, Guide signage, Pedestrian signage, Traffic light, Traffic signal, Street lamp, Traffic sign, Warning sign, Compulsory sign, Regulatory sign, Informatory sign, Construction sign, Over-road gantry sign panels, Barrier, Guardrail, Bollard, Billboard, Traffic cone, Construction cone, Sidewalk, Street light, Directional Highway, Motorway Gantry Signage, Modular utility building. \\
    \midrule
    others &
    Pedestrian, Broken yellow line, Double solid yellow line, Solid continuous yellow line, Broken white line, Double solid white line, Solid continuous white line, Dotted white line, Broken line, Broken line and solid line, Double solid line, Single yellow line, Single white line, Center line, Lane line, Edge line. \\
    \bottomrule
  \end{tabular}
  \caption{OWL-ViT generic vocabulary elements used for the anomaly detection.}
  \label{tab:my-table}
\end{table*}

\subsection{Real edge-case queries}
For the real edge-case queries, a unified anomaly vocabulary of text phrases describing specific semantic anomalies was also introduced to the object detector, as listed in Table \ref{tab:edgeinput}. These queries correspond to the 12 real-world failure scenarios detailed in Table \ref{tab1}. Each was formulated using descriptive predicates to reflect the contextual setting (e.g. \textit{"on a road", "across a roadway",} etc.). We then passed the inputs to the pre-trained OWL-ViT model (forward pass) to get object detection predictions. For each image, we obtained the detected object labels, their bounding boxes and confidence scores for each detection.
These outputs were filtered using a separate confidence thresholds:

• normal threshold: only retains predictions for the normal classes with confidence > threshold.

• anomaly threshold: only retains predictions for the anomalous classes with confidence > threshold.

The selection of two different thresholds, reflects different priors: normal objects must be detected reliably, whereas rare anomalies may be meaningful even with lower confidence.  

The output predictions were based on the resized input image. This is because OWL-ViT outputs normalized box coordinates in [cx, cy, w, h] format assuming a fixed input image size. Therefore, we converted the model outputs to a COCO API format and retrieve rescaled coordinates (with respect to the original image sizes) in [x0, y0, x1, y1] format.

\begin{table*}[t]
  \centering
  \small
  \begin{tabular}{p{4.5cm} p{11.5cm}}
    \toprule
    \textbf{Vocabulary type} & \textbf{Vocabulary objects} \\
    \midrule
    edge\_cases &
    ``rear of a vehicle with an advertisement depicting cyclists riding on a road'', \newline
    ``rear of a vehicle covered with a printed poster depicting a mountainous landscape'', \newline
    ``a series of parallel continuous thin stripes'', \newline
    ``rear of a truck with printed graphics resembling stop signs'', \newline
    ``maintenance truck carrying portable traffic lights'', \newline
    ``realistic-looking road backdrop printed on a large panel positioned across a roadway'', \newline
    ``a vertical white promotional flag banner'', \newline
    ``a photo of a road blocked by railway crossing while train passes'', \newline
    ``road parapet gradually encroached into the driving lane'', \newline
    ``a buckboard wagon traveling on a public road alongside vehicles'', \newline
    ``road maintenance workers operating water tanker trucks spraying water across a road'', \newline
    ``a traffic officer is manually directing vehicles in an urban setting'' \\
    \bottomrule
  \end{tabular}
  \caption{Common OWL-ViT edge case vocabulary elements used for the real edge cases.}
  \label{tab:edgeinput}
\end{table*}

Finally, we visualized the predictions on the input image. To distinguish between normal and anomalous predictions, we applied color-coding scheme: green bounding boxes indicate objects from the normal class vocabulary, while red bounding boxes highlight detections associated with the anomaly queries. This visualization allowed for both qualitative analysis of the model's outputs and intuitive understanding of the spatial context of anomalies within the scene.

The final predictions of the model were used to generate a scene description for each image. After carefully thresholding, tuning and filtering, each retained detection was transformed into a natural language phrase describing the identified object or semantic element. For normal classes, this often resulted in simple object-centric descriptions (e.g. \textit{"a tree", "a car", "a traffic sign",} etc.). However, in cases involving edge cases, the output frequently captured a more holistic interpretation of the scene (e.g. \textit{"a series of parallel continuous thin stripes", "a maintenance truck carrying portable traffic lights", "rear of a vehicle with an advertisement depicting cyclists riding on a road",} etc.). 

\subsection{LLM anomaly detection pipeline}

\begin{figure}[t]
  \centering
   \includegraphics[width=0.8\linewidth]{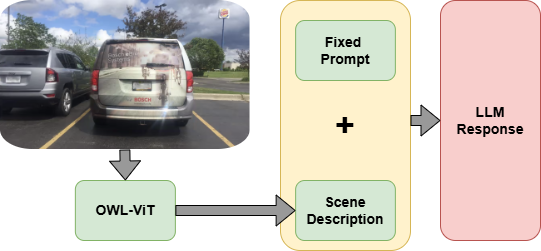}
   \caption{The proposed framework pipeline.}
    \label{fig:pipeline}
\end{figure}

These descriptions reflected not just individual entities but the semantic character of the entire image when the anomaly was scene-level. This context-rich representation formed the basis of the prompt passed to the LLM, effectively converting the visual content of the image into a structured textual summary suitable for high level reasoning. 

The reason we selected to leverage LLMs as a “semantic reasoning” module that monitors the AV observations instead of existing methods for anomaly detection, is because existing approaches, cannot reason about AV failure modes that arise from semantically anomalous scenarios. Instead, they only aim to detect anomalies correlated with inference errors or to measure how uncertain the perception system and its outputs are. In addition, they cannot be deployed as off-the-shelf diagnostic tools because they typically require access to model's training data, annotated failure examples or extensive domain expertise \cite{elhafsi2023semantic}.

The resulted scene description was then integrated into a custom prompt template that incorporates chain-of-thought prompting to identify whether any element of the scene might cause erroneous, unsafe, or unexpected behavior. Chain-of-thought reasoning adds task and system specific structure to the reasoning. 

Large Language models draw on a wide-ranging training data that gives them broad reasoning skills, yet any deployed robotic system operates with its own unique abilities and limitations. To embed an LLM-based anomaly monitor inside an autonomy task, the model must be aware of those platform-specific strengths and constraints \cite{elhafsi2023semantic}. For instance, a self-driving car calibrated for congested streets will exhibit different failure patterns from one optimized for rural highways. To an extent, this is achieved through chain-of-thought prompting, as the logical structure “prime” the LLM as to how to characterize anomalies. 

Then, we fed the prompt to several state-of-the-art LLMs, each of which analyzes it and identifies any potential semantic anomalies present in the observation, as shown in Fig. \ref{fig:pipeline}. The models were tasked with classifying each scenario as either \textit{Normal} or \textit{Anomaly} and to assign a confidence score reflecting their certainty in classification. 

\section{Experimental Results}

Table \ref{tab:results} summarizes the anomaly classifications and percentage confidence scores, produced by each LLM across the 12 edge cases. As shown, all models were able to identify anomalies in several scenarios, however, the consistency and confidence levels varied both across models and among edge cases. For most edge cases, the majority of the models classified the scene as an anomaly, in alignment with how a human driver would recognize it. 

Across all models, \textit{'Meta-Llama-3.1-8B-Instruct-Turbo'} consistently achieved the highest anomaly detection confidence, outperforming the others, in 7 out of 12 cases. Its judgments tended to align with scene complexity and risk implications, showing strong alignment with human reasoning, particularly in edge cases involving visual deception (e.g. realistic stop signs images or misleading advertisements posters). For example, in Edge Case 6, where printed stop sign graphics on a truck misled perception systems, the model classified the anomaly with $95\%$ confidence, correctly recognizing the contextual misplacement.

\textit{'Mixtral-8x7B-Instruct-v0.1'} also demostrated strong and consistent performance, especially in high-risk perception cases (Edge Cases 2, 3, 4, 5, 10, and 11). It classified every scenario as anomalous and maintained confidence levels above $80\%$ in most, suggesting a more conservative anomaly bias. However in low-ambiguity cases such as Edge Cases 1 and 7, it produced relatively low confidence scores ($5\%$), indicating possible difficulties with scene descriptions containing ambiguous or low-frequency features.  

On the other hand, \textit{'Qwen2.5-7B-Instruct-Turbo'} shown the lowest confidence in its classifications and the greatest variability, producing both high and extremely low confidence scores ($5\%$ and $95\%$). While it successfully detected anomalies in several complex scenes, its judgments were inconsistent and sometimes counterintuitive (e.g. assigning only $10\%$ confidence score to Edge Case 1 despite the presence of visual illusions affecting perception). This suggests sensitivity to phrasing nuances or a shallower reasoning trajectory, even under consistent prompt formatting.

Finally, \textit{'Nvidia-Llama-3.1-Nemotron-70B-Instruct-HF'} model occupied a middle ground, performing reliably in the majority of cases but with moderate confidence levels. It demonstrated notable robustness in harder edge cases (e.g. Edge Case 8, with $92\%$ confidence), but was less confident than Meta-Llama or Mixtral on average. This variability underscores the impact of the model's properties (e.g. \textit{"architecture", "number of parameters", "training data","tuning methods",} etc) in its classification properties, despite the identical scene description and the prompt template.

The text queries used in OWL-ViT, which depend heavily on precise text prompting and appropriate threshold tuning, combined both a comprehended vocabulary for common road objects and a set of carefully described edge-case scenarios. As every zero-shot object detector, OWL-ViT is sensitive to the composition and phrasing of the text queries used during inference, as well. In addition, prompting techniques such as ``chain-of-thought reasoning'', affect of how the LLM reasons, therefore, understanding the semantic context and classifying correctly each road object the autonomous vehicle observes. Finally, due to the reason that our dataset consisted of real-world images the object detector would occasionally misclassify objects or hallucinate unseen objects altogether, possibly due to lighting conditions, blurring and camera noise.

\begin{table*}[t]
  \centering
  \footnotesize
  \begin{tabular}{T{0.8cm} C{2.2cm} T{12.5cm}}
    \toprule
    \textbf{Edge Case \#} & \textbf{Resulted image} &
    \textbf{LLM response} \\
    \midrule
    1 & \includegraphics[width=\linewidth,height=1.5cm,keepaspectratio]{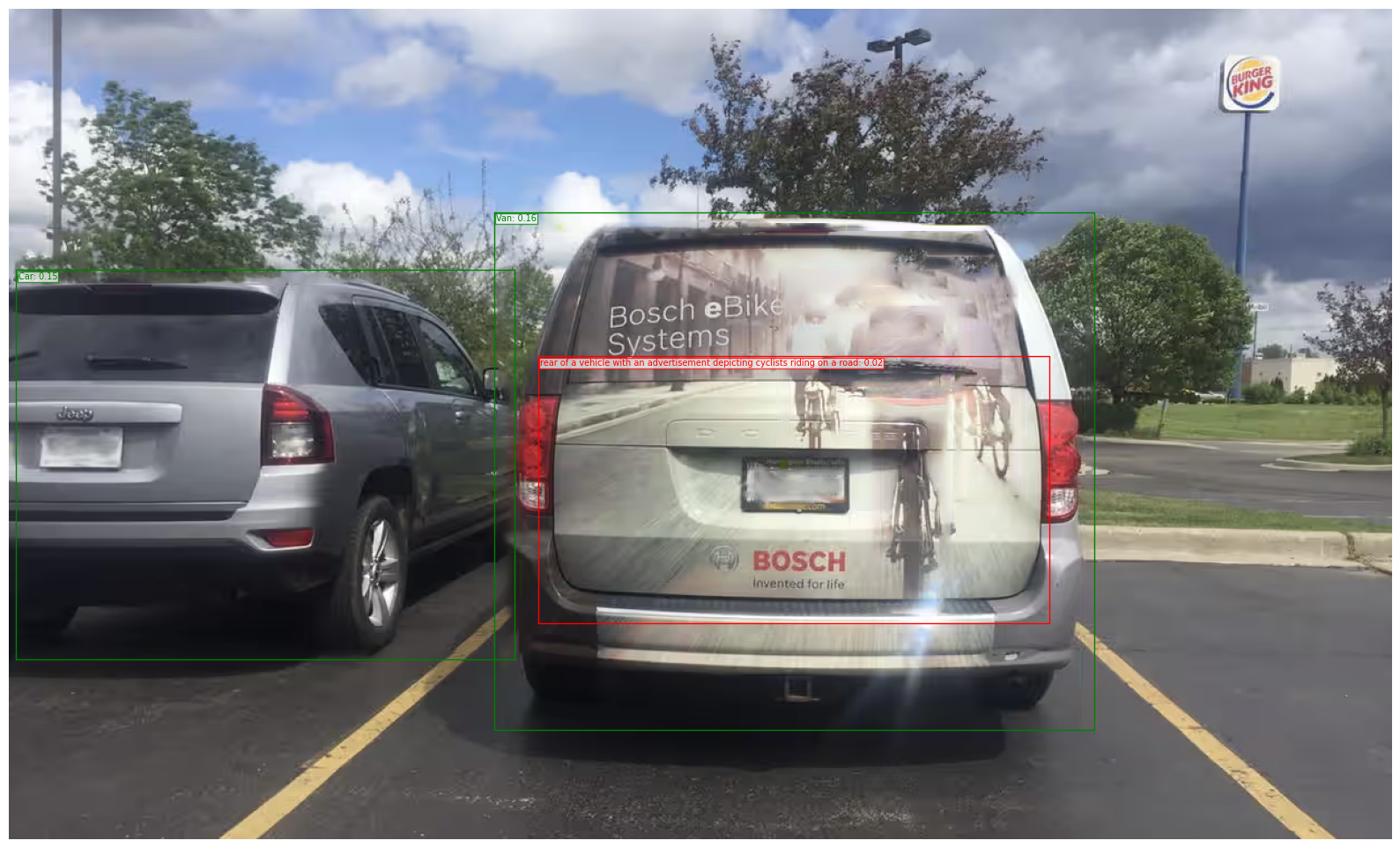} &
    \footnotesize{Model \textit{Mixtral-8x7B-Instruct-v0.1} 
    - Anomaly Confidence: 5\%

    Model \textit{Qwen2.5-7B-Instruct-Turbo}
    - Anomaly Confidence: 10\%

    Model \textit{Nvidia-Llama-3.1-Nemotron-70B-Instruct-HF}
    - Anomaly Confidence: 6\%

    Model \textit{Meta-Llama-3.1-8B-Instruct-Turbo}
    - Anomaly Confidence: \textbf{90\%}}
    \\ \hline
    
    \midrule
    2 & \includegraphics[width=\linewidth,height=1.5cm,keepaspectratio]{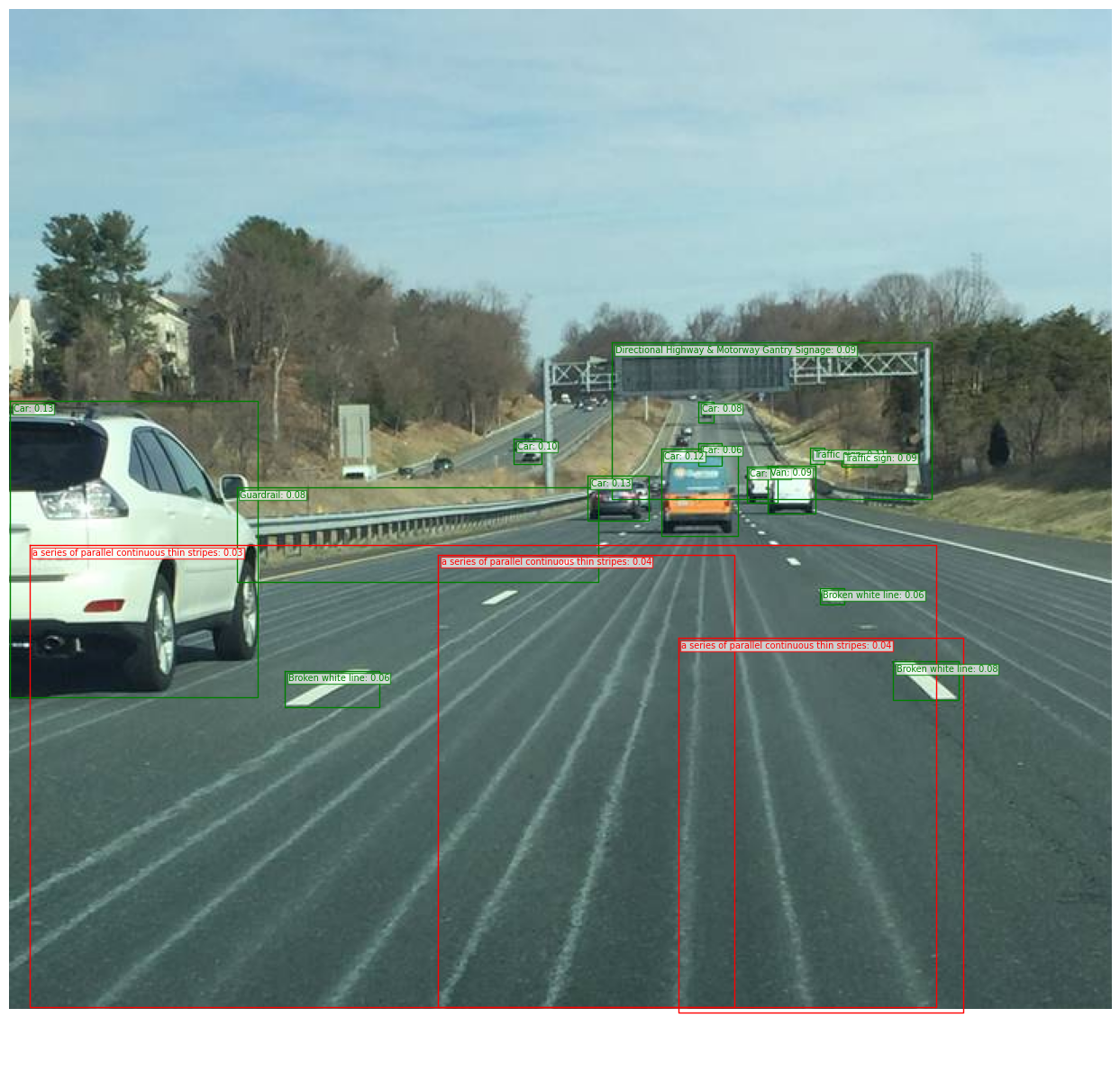} &
    \footnotesize{Model \textit{Mixtral-8x7B-Instruct-v0.1} 
    - Anomaly Confidence: 85\%
    
    Model \textit{Qwen2.5-7B-Instruct-Turbo} 
    - Anomaly Confidence: 15\%
    
    Model \textit{Nvidia-Llama-3.1-Nemotron-70B-Instruct-HF} 
    - Anomaly Confidence: 70\%
    
    Model \textit{Meta-Llama-3.1-8B-Instruct-Turbo}
    - Anomaly Confidence: \textbf{85.71\%}}
    \\ \hline
    
    \midrule
    3 & \includegraphics[width=\linewidth,height=1.5cm,keepaspectratio]{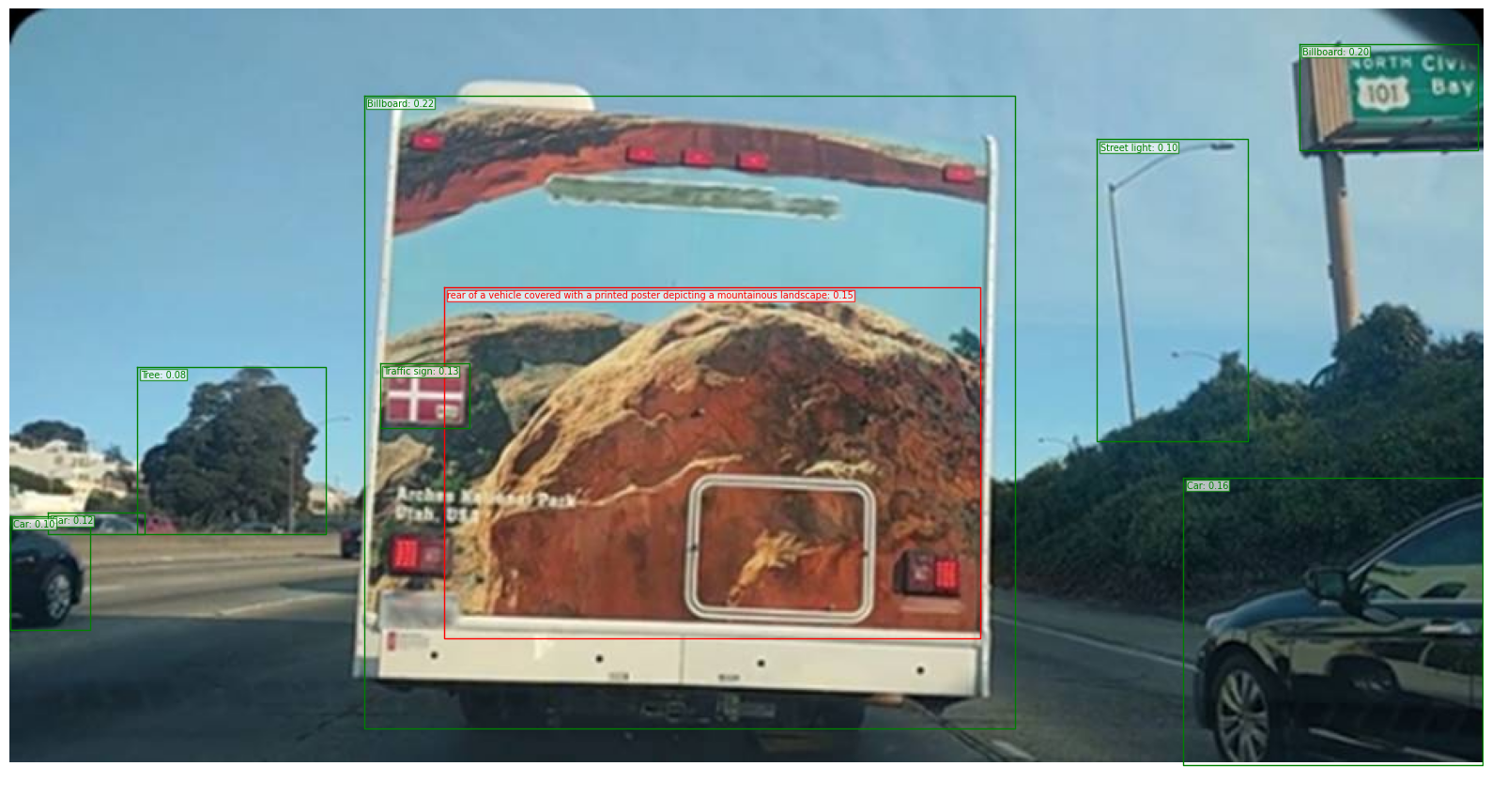} &
    \footnotesize{Model \textit{Mixtral-8x7B-Instruct-v0.1} 
    - Anomaly Confidence: \textbf{85\%}
    
    Model \textit{Qwen2.5-7B-Instruct-Turbo}
    - Anomaly Confidence: 83.33\%
    
    Model \textit{Nvidia-Llama-3.1-Nemotron-70B-Instruct-HF} 
    - Anomaly Confidence: 60\%
    
    Model \textit{Meta-Llama-3.1-8B-Instruct-Turbo}
    - Anomaly Confidence: 83\%}
    \\ \hline
    
    \midrule
    4 & \includegraphics[width=\linewidth,height=1.5cm,keepaspectratio]{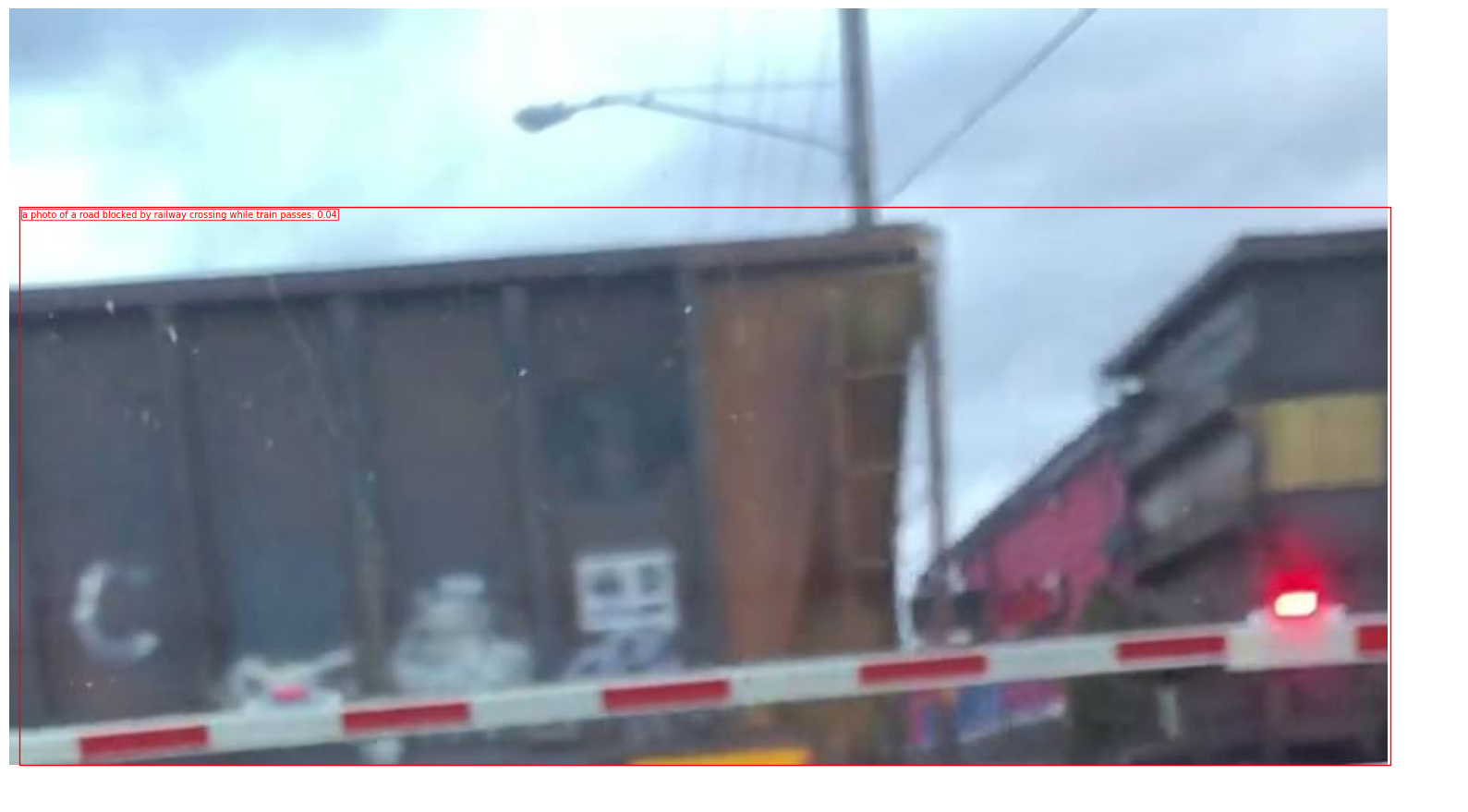} &
    \footnotesize{Model \textit{Mixtral-8x7B-Instruct-v0.1}
    - Anomaly Confidence: 90\%
    
    Model \textit{Qwen2.5-7B-Instruct-Turbo}
    - Anomaly Confidence: \textbf{95\%}
    
    Model \textit{Nvidia-Llama-3.1-Nemotron-70B-Instruct-HF}
    - Anomaly Confidence: 80\%
    
    Model \textit{Meta-Llama-3.1-8B-Instruct-Turbo}
    - Anomaly Confidence: \textbf{95\%}}
    \\ \hline
    
    \midrule
    5 & \includegraphics[width=\linewidth,height=1.5cm,keepaspectratio]{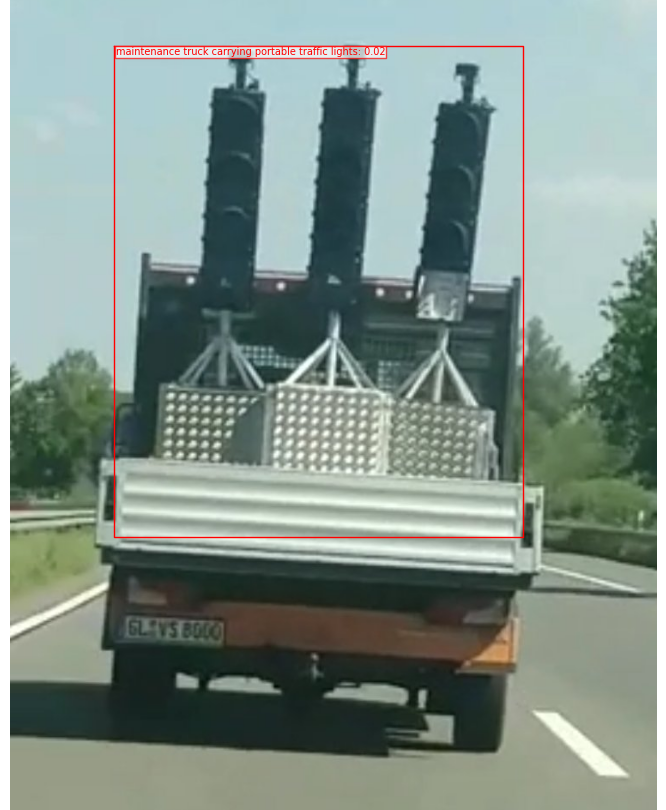} &
    \footnotesize{Model \textit{Mixtral-8x7B-Instruct-v0.1}
    - Anomaly Confidence: \textbf{85\%}
    
    Model \textit{Qwen2.5-7B-Instruct-Turbo}
    - Anomaly Confidence: 60\%
    
    Model \textit{Nvidia-Llama-3.1-Nemotron-70B-Instruct-HF}
    - Anomaly Confidence: 18\%
    
    Model \textit{Meta-Llama-3.1-8B-Instruct-Turbo}
    - Anomaly Confidence: \textbf{85\%}}
    \\ \hline
    
    \midrule
    6 & \includegraphics[width=\linewidth,height=1.5cm,keepaspectratio]{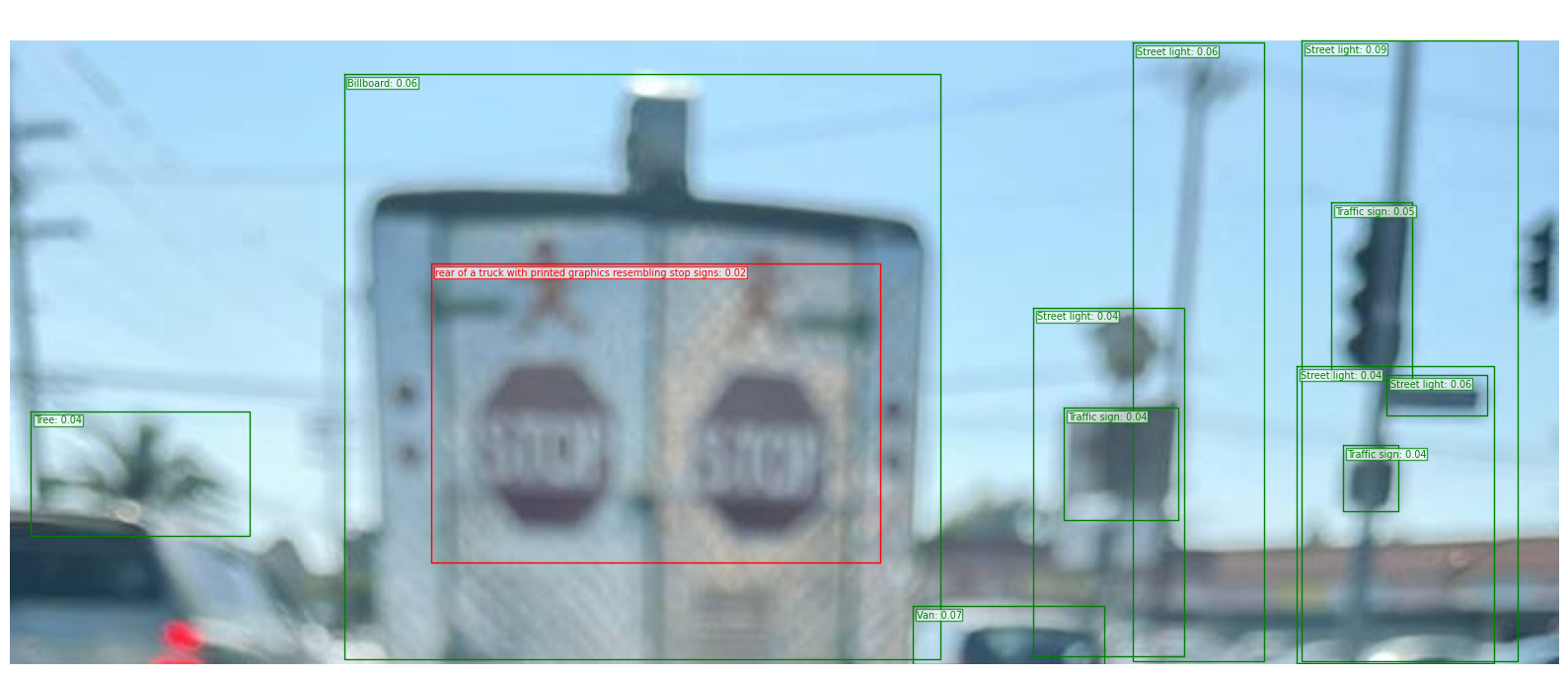} &
    \footnotesize{Model \textit{Mixtral-8x7B-Instruct-v0.1}
    - Anomaly Confidence: 90\%
    
    Model \textit{Qwen2.5-7B-Instruct-Turbo}
    - Anomaly Confidence: 14.28\%
    
    Model \textit{Nvidia-Llama-3.1-Nemotron-70B-Instruct-HF}
    - Anomaly Confidence: 45\%
    
    Model \textit{Meta-Llama-3.1-8B-Instruct-Turbo}
    - Anomaly Confidence: \textbf{95\%}}
    \\ \hline
    
    \midrule
    7 & \includegraphics[width=\linewidth,height=1.4cm,keepaspectratio]{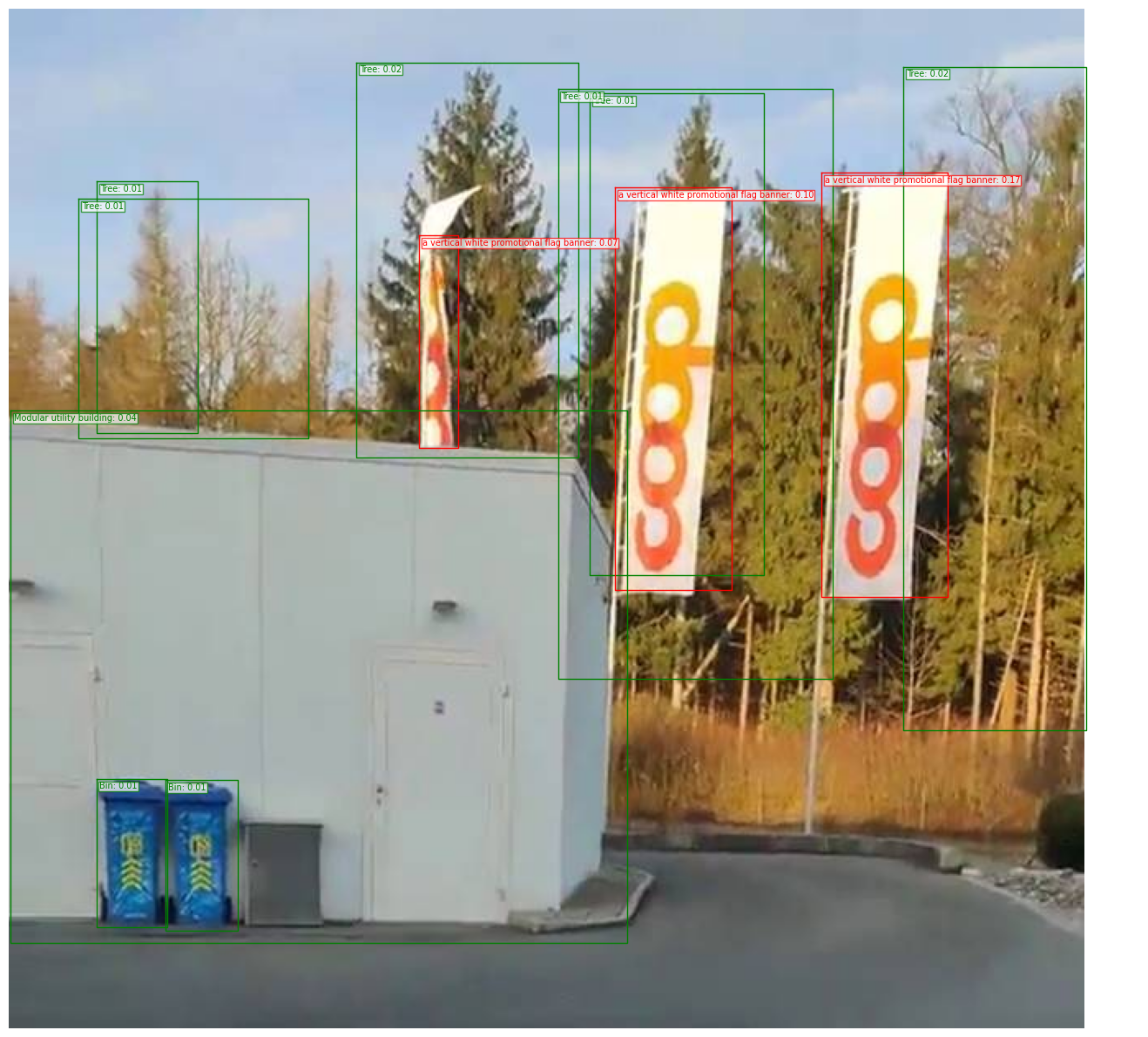} &
    \footnotesize{Model \textit{Mixtral-8x7B-Instruct-v0.1}
    - Anomaly Confidence: 5\%
    
    Model \textit{Qwen2.5-7B-Instruct-Turbo}
    - Anomaly Confidence: 66.67\%
    
    Model \textit{Nvidia-Llama-3.1-Nemotron-70B-Instruct-HF}
    - Anomaly Confidence: 60\%
    
    Model \textit{Meta-Llama-3.1-8B-Instruct-Turbo}
    - Anomaly Confidence: \textbf{80\%}}
    \\ \hline
    
    \midrule
    8 & \includegraphics[width=\linewidth,height=1.5cm,keepaspectratio]{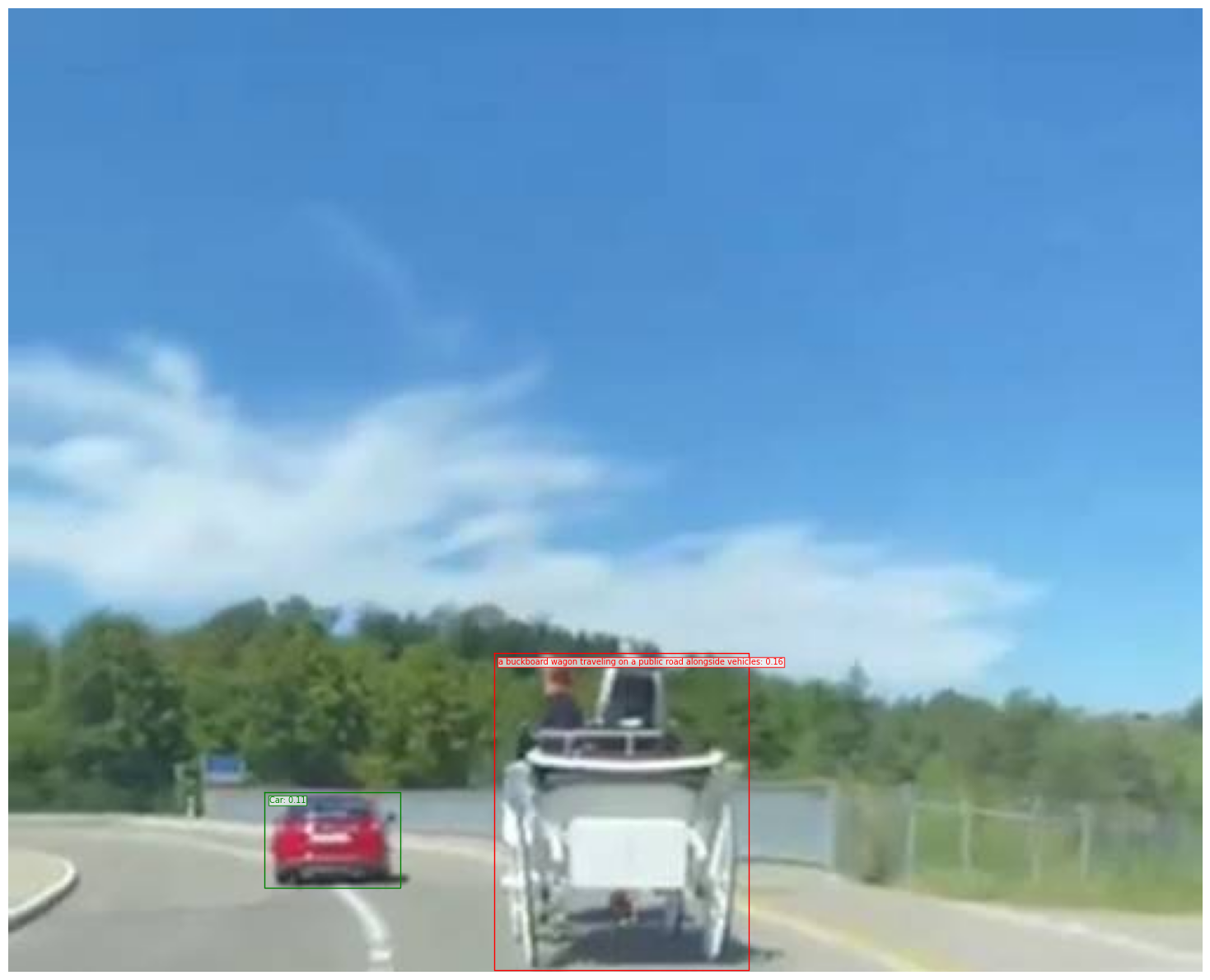} &
    \footnotesize{Model \textit{Mixtral-8x7B-Instruct-v0.1}
    - Anomaly Confidence: 80\%
    
    Model \textit{Qwen2.5-7B-Instruct-Turbo}
    - Anomaly Confidence: 75\%
    
    Model \textit{Nvidia-Llama-3.1-Nemotron-70B-Instruct-HF}
    - Anomaly Confidence: \textbf{92\%}
    
    Model \textit{Meta-Llama-3.1-8B-Instruct-Turbo}
    - Anomaly Confidence: 75\%}
    \\ \hline
    
    \midrule
    9 & \includegraphics[width=\linewidth,height=1.5cm,keepaspectratio]{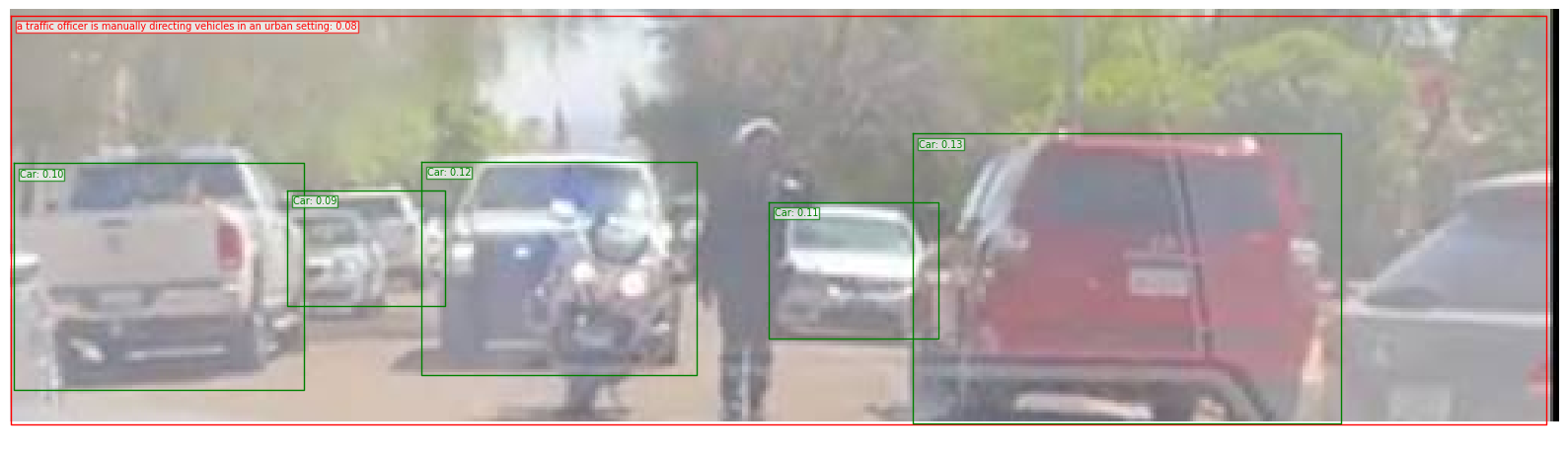} &
    \footnotesize{Model \textit{Mixtral-8x7B-Instruct-v0.1}
    - Anomaly Confidence: \textbf{85\%}
    
    Model \textit{Qwen2.5-7B-Instruct-Turbo}
    - Anomaly Confidence: 5\%
    
    Model \textit{Nvidia-Llama-3.1-Nemotron-70B-Instruct-HF}
    - Anomaly Confidence: 82\%
    
    Model \textit{Meta-Llama-3.1-8B-Instruct-Turbo}
    - Anomaly Confidence: 10\%}
    \\ \hline
    
    \midrule
    10 & \includegraphics[width=\linewidth,height=1.5cm,keepaspectratio]{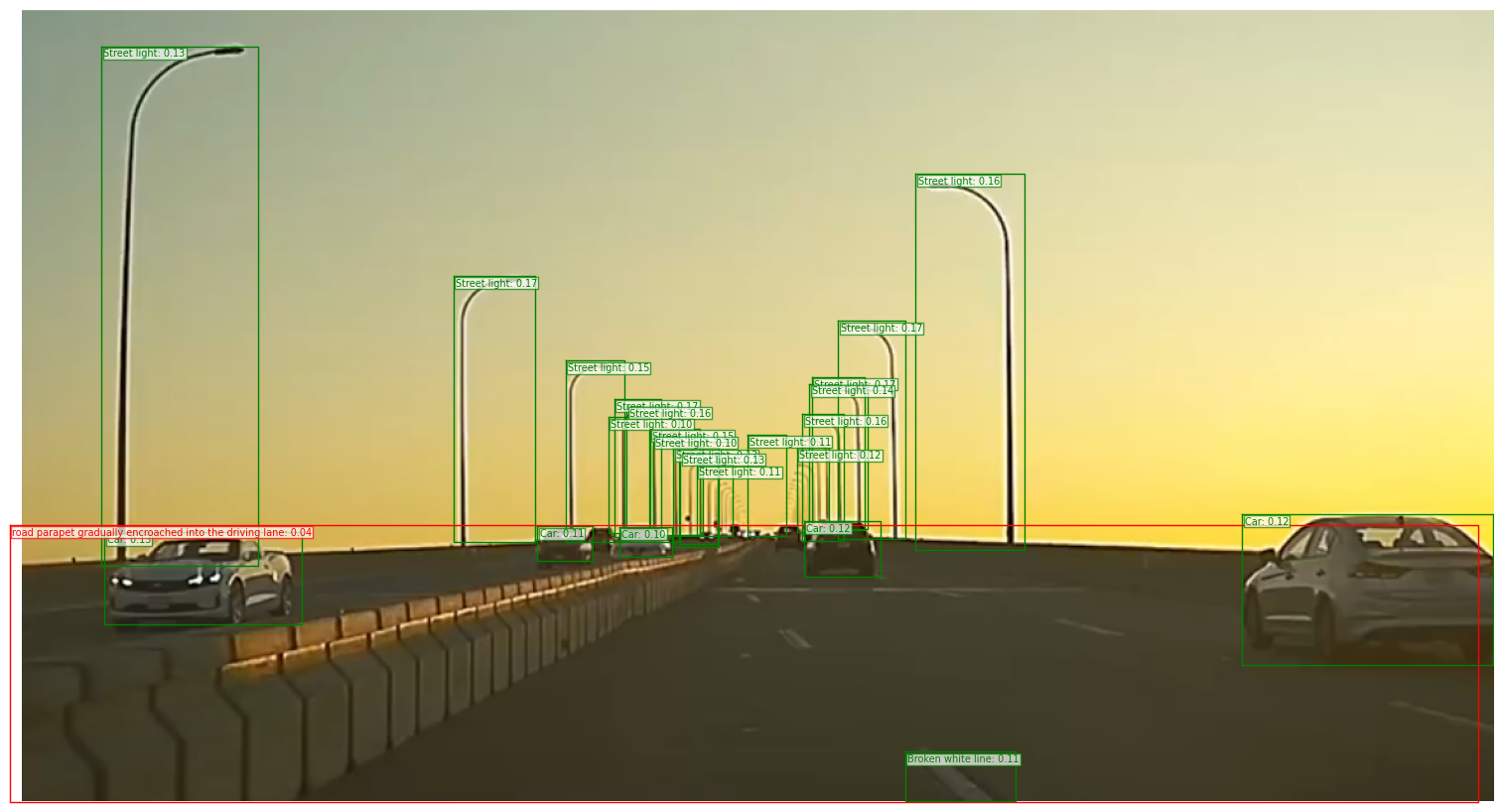} &
    \footnotesize{Model \textit{Mixtral-8x7B-Instruct-v0.1}
    - Anomaly Confidence: \textbf{90\%}
    
    Model \textit{Qwen2.5-7B-Instruct-Turbo}
    - Anomaly Confidence: 75\%
    
    Model \textit{Nvidia-Llama-3.1-Nemotron-70B-Instruct-HF}
    - Anomaly Confidence: 88\%
    
    Model \textit{Meta-Llama-3.1-8B-Instruct-Turbo}
    - Anomaly Confidence: 80\%}
    \\ \hline
    
    \midrule
    11 & \includegraphics[width=\linewidth,height=1.5cm,keepaspectratio]{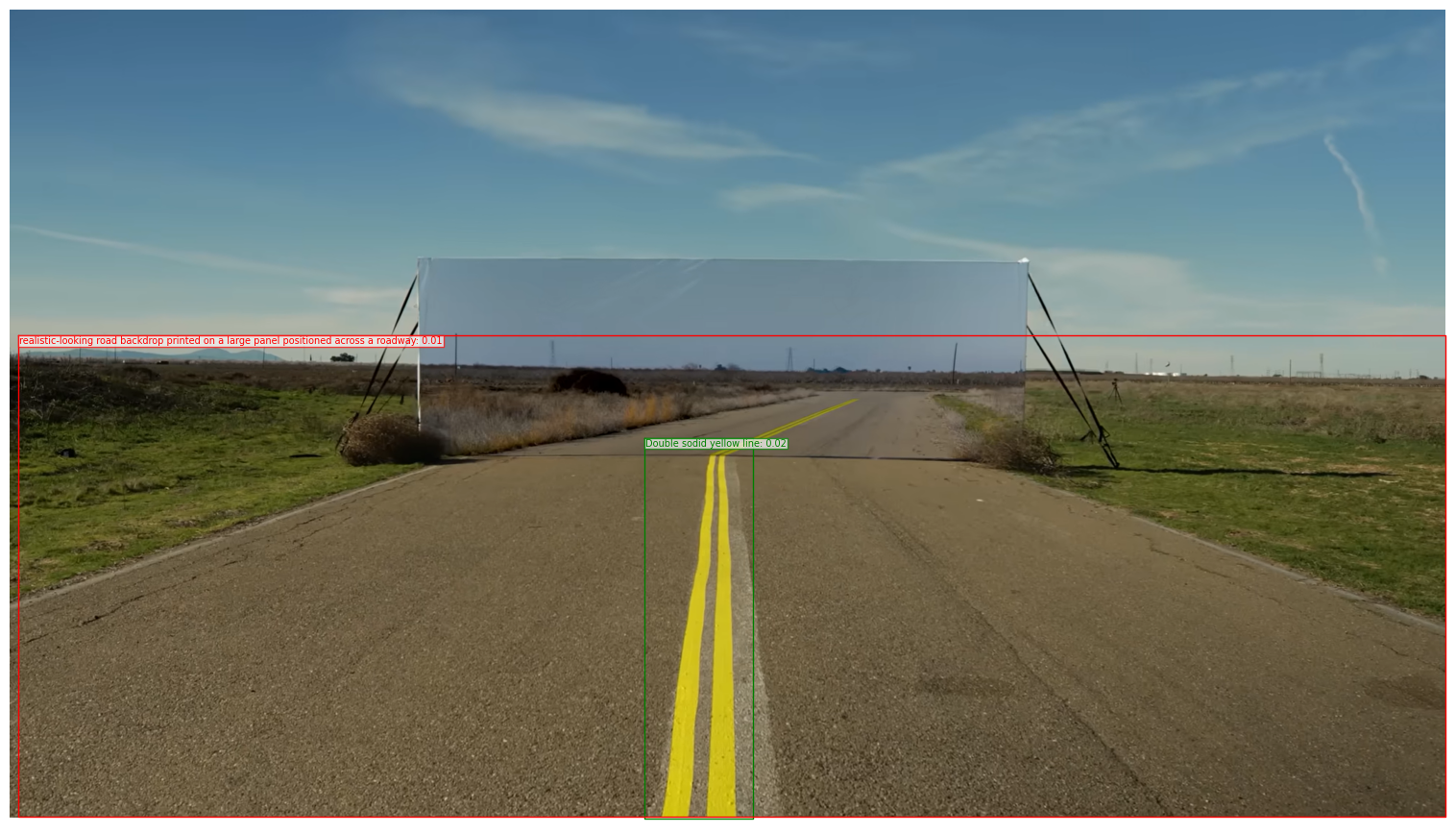} &
    \footnotesize{Model \textit{Mixtral-8x7B-Instruct-v0.1} 
    - Anomaly Confidence: 90\%
    
    Model \textit{Qwen2.5-7B-Instruct-Turbo}
    - Anomaly Confidence: 90\%
    
    Model \textit{Nvidia-Llama-3.1-Nemotron-70B-Instruct-HF} 
    - Anomaly Confidence: 85\%
    
    Model \textit{Meta-Llama-3.1-8B-Instruct-Turbo}
    - Anomaly Confidence: \textbf{95\%}}
    \\ \hline
    
    \midrule
    12 & \includegraphics[width=\linewidth,height=1.5cm,keepaspectratio]{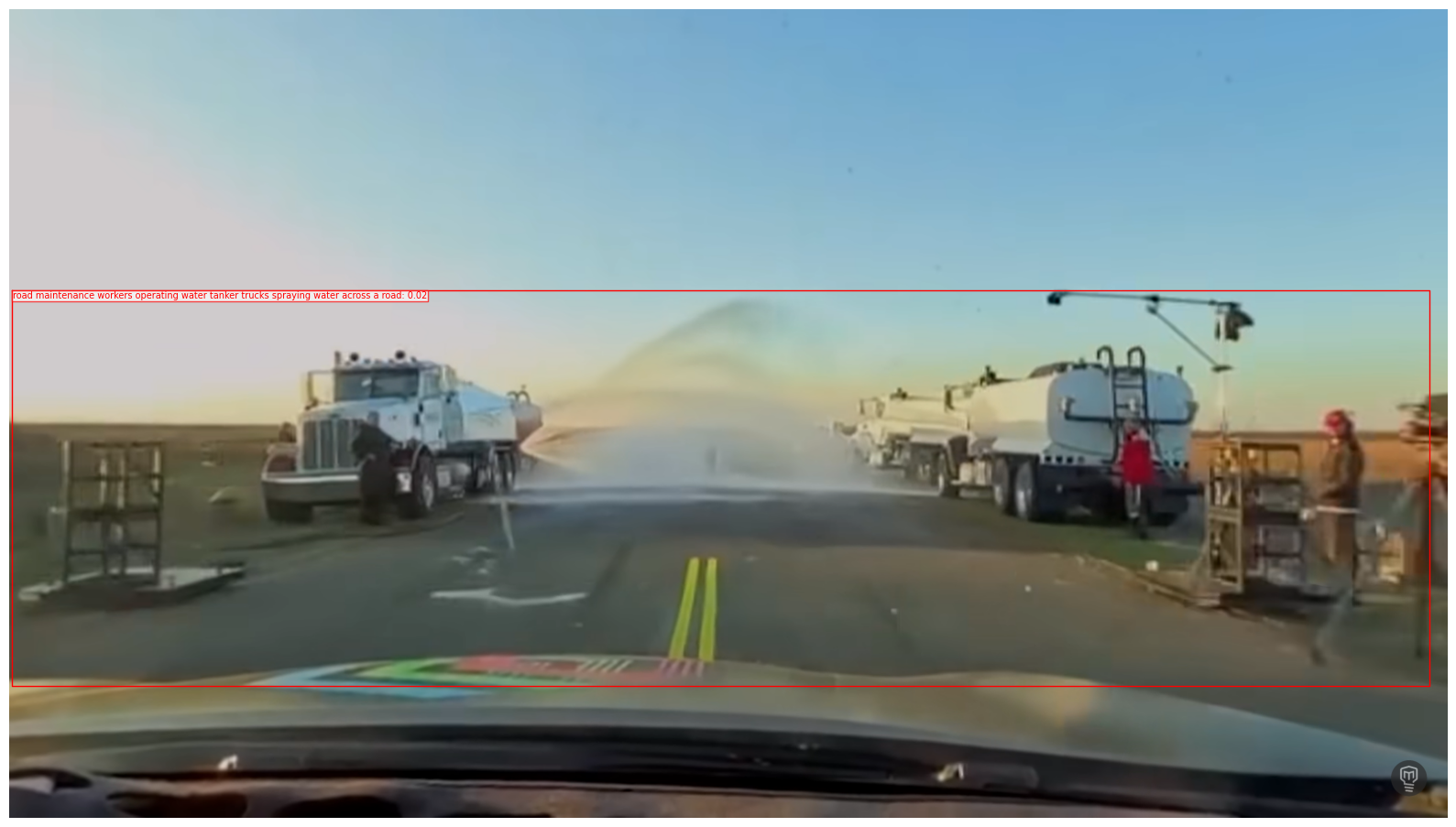} &
    \footnotesize{Model \textit{Mixtral-8x7B-Instruct-v0.1}
    - Anomaly Confidence: 85\%
    
    Model \textit{Qwen2.5-7B-Instruct-Turbo}
    - Anomaly Confidence: 70\%
    
    Model \textit{Nvidia-Llama-3.1-Nemotron-70B-Instruct-HF}
    - Anomaly Confidence: \textbf{92\%}
    
    Model \textit{Meta-Llama-3.1-8B-Instruct-Turbo}
    - Anomaly Confidence: 85\%}
    \\ \hline
    
    \bottomrule
  \end{tabular}
  \caption{Resulted scene descriptions}
  \label{tab:results}
\end{table*}

\section{Conclusions}

By leveraging a hand-curated dataset of real-world AV edge cases and systematically comparing state-of-the-art LLMs, we have demonstrated the potential and the current limitations of LLM-based semantic anomaly detectors when applied to safety-critical driving scenarios.

Our results, highlight three different key insights. First, the construction of the visual scene description-including vocabulary coverage, query phrasing, and object specificity-plays a critical role in detection success. Second, LLM performance varies significantly across models, suggesting that architectural differences, training data, and tuning strategies, all impact the reliability of reasoning-based anomaly detection. Finally, we showed that prompting techniques, such as chain-of-thought reasoning, enhance the interpretability and granularity of LLM decisions, offering transparent justification for each classification.

Ultimately, this research demonstrates the feasibility and value of incorporating LLMs as contextual monitors within AV systems. Rather than replacing existing components, LLMs can serve as complementary semantic reasoning agents, particularly in situations where visual ambiguity, contextual deception, or rare configurations may compromise safety. 

Future work includes i) integration of multi-modal LLMs, capable of directly processing visual data alongside text, ii) the development of prompt strategies or/and templates tailored to specific AV stacks, and iii) the synergistic combination of LLM monitors with established out-of-distribution (OOD) and detection methods. By advancing these avenues, LLM-based anomaly detection may become a more robust, human-aligned component within the broader safety and reliability framework for next-generation autonomous vehicles.

{\small
\bibliographystyle{ieee_fullname}
\bibliography{main}
}

\end{document}